\documentclass{article}
\usepackage{PRIMEarxiv}
\usepackage[utf8]{inputenc} % allow utf-8 input
\usepackage[T1]{fontenc}    % use 8-bit T1 fonts
\usepackage{hyperref}       % hyperlinks
\usepackage{url}            % simple URL typesetting
\usepackage{booktabs}       % professional-quality tables
\usepackage{amsfonts}       % blackboard math symbols
\usepackage{nicefrac}       % compact symbols for 1/2, etc.
\usepackage{microtype}      % microtypography
\usepackage{lipsum}
\usepackage{fancyhdr}       % header
\usepackage{graphicx}       % graphics
\graphicspath{{media/}}     % organize your images and other figures under media/ folder
\usepackage{amsmath}
\usepackage{hyperref}
\hypersetup{colorlinks,citecolor=blue, linkcolor=red}
\usepackage{orcidlink}
\title{Panorama Tomosynthesis from Head CBCT\\
with Simulated Projection Geometry
}

\author{
  Anusree P.S. \orcidlink{0009-0006-9381-3618} \textsuperscript{1}, Bikram Keshari Parida \orcidlink{0000-0003-1204-357X}\textsuperscript{1}, Seong Yong Moon\orcidlink{0000-0002-7513-4404}\textsuperscript{2}, Wonsang You \orcidlink{0000-0002-6806-7135} \textsuperscript{1,*} \\
  \textsuperscript{1} AIIP Lab, Sun Moon University, Asan, South Korea.\\
  \textsuperscript{2} College of Dentistry, Chosun University, Gwangju, South Korea. \\
  \texttt{\{anusreepandath, parida.bikram90.bkp\}@gmail.com, msygood@chosun.ac.kr}  \\
  \textsuperscript{*}corresponding author: \texttt{wyou@kaist.ac.kr}  \\
}

\begin{document}
\maketitle

\begin{abstract}
% Cone Beam Computed Tomography(CBCT) and Panoramic X-rays are the most commonly used imaging modalities in dental health care. Dentists mostly prefer CBCT for surgical planning because of its capability to produce three-dimensional views of a patient's head. Despite its advantages, additional X-rays may be required to analyze the teeth and jaw demography. A Panoramic X-ray is beneficial in such a situation since it can capture the entire maxillofacial region in a single image. Nevertheless, it is advantageous to synthesize a panoramic view from the current CBCT volume in cases where a panoramic X-ray is unavailable or when the patient cannot undergo an immediate additional scan. For X-ray synthesis, most of the earlier works required the extraction of the patient-specific dental arch. Also, the quality of reconstructed X-rays varies depending on the CBCT acquisition parameters and patient positioning. So, we aim to synthesize Panoramic X-rays from diverse head CBCTs based on a simulated projection geometry with moving rotation centres, thereby eliminating the need for dental arch extraction. 

Cone Beam Computed Tomography (CBCT) and Panoramic X-rays are the most commonly used imaging modalities in dental health care. CBCT can produce three-dimensional views of a patient's head, providing clinicians with better diagnostic capability, whereas Panoramic X-ray can capture the entire maxillofacial region in a single image. If the CBCT is already available, it can be beneficial to synthesize a Panoramic X-ray, thereby avoiding an immediate additional scan and extra radiation exposure. Existing methods focus on delineating an approximate dental arch and creating orthogonal projections along this arch. However, no golden standard is available for such dental arch extractions, and this choice can affect the quality of synthesized X-rays. To avoid such issues, we propose a novel method for synthesizing Panoramic X-rays from diverse head CBCTs, employing a simulated projection geometry and dynamic rotation centers.  Our method effectively synthesized panoramic views from CBCT, even for patients with missing or nonexistent teeth and in the presence of severe metal implants. Our results demonstrate that this method can generate high-quality panoramic images irrespective of the CBCT scanner geometry.
\end{abstract}

% keywords can be removed

\section{Introduction}

Dentistry witnessed rapid advancements in its diagnostic imaging techniques, providing clinicians with comprehensive and detailed information for accurate treatment planning. The emergence of CBCT systems with panoramic and cephalometric scanning capability and the design of panoramic systems with auto-focusing functionality are the aftereffects of such innovations and research endeavors. Despite the wide range of imaging methods available \cite{white_2014}, extraoral methods like CBCTs and panoramic X-rays \cite{suomalainen_2015} have seen a sharp increase in demand because they are more convenient for patients to use during imaging. Panoramic X-rays, known as orthopantomograms (OPGs), have long been a staple in dental imaging due to their ability to capture a full view of the entire maxillofacial region in a single image with low radiation exposure to patients \cite{farman_2007}. In contrast, CBCTs can offer sharp, detailed, three-dimensional views of the head, but at the expense of increased radiation exposure \cite{pauwels_2015, al-okshi_2013, ludlow_2015}. CBCT has demonstrated its efficacy across diverse applications, encompassing disease prognosis and bone density analysis \cite{hung_2022, lee_2016}. The extraction of mandibular shape \cite{hwang_2018, zeller_2020} and estimation of dental arches \cite{bae_2019, zhu_2021} represent pivotal stages in the analysis and visualization of CBCT data. The efficacy and quality of CBCT are intrinsically linked to the scanner parameters and the acquisition and projection geometries \cite{icen_2020, shokri_2018}. Conversely, panoramic X-rays are frequently afflicted by challenges such as ghostly shadows, distortions, and the superimposition of bony and dental structures. Additionally, inaccuracies can arise due to patient misalignment, resulting in disproportionate teeth and jaw. Moreover, errors in measurements and assessments may emanate from image distortions induced by the acquisition geometry, potentially leading to misinterpretations in diagnostic outcomes \cite{izzetti_2021}. Scholars have systematically examined and contrasted the accuracy of panoramic radiographs and CBCT concerning applications tailored to specific tasks \cite{lim_2018, shahidi_2018, tang_2017, lopes_2016}.

The examining physician determines the imaging modality based on the diagnostic task and the patient's condition. While most tasks can resort to panoramic X-rays with comparable precision, they are not a suitable substitute for CBCT in some situations, such as abnormal tooth visualization, surgical planning, and implant treatment \cite{ozalp_2018, jacobs_2018}. To avoid unnecessary scanning, dentists prefer to use CBCTs as their primary imaging technique rather than panoramic X-rays. However, subsequent phases of the diagnostic process may necessitate the utilization of two-dimensional X-ray imaging. Due to the augmented radiation exposure, subjecting patients to additional scans promptly following CBCT is impractical. Furthermore, preparing uncooperative pediatric and adult patients for subsequent scans poses considerable challenges. In such scenarios, the preferable course of action would involve the synthesis of a panoramic X-ray utilizing the existing CBCT data. While this approach may not serve as a substitute for the authentic X-ray, it serves the purpose of delaying the necessity for additional scans. This delay is especially beneficial to mitigate the potential adverse effects of repeated radiation exposure on the patient's well-being. Post-processing and visualization methods using CBCT, such as Curved Multi-Planar Reformatting (CMPR) and Direct Volume Rendering (DVR), facilitate the generation of panoramic views by jaw reformatting. Besides the CBCT reconstruction software, third-party applications are also available that can offer functionalities to manually outline a focal plane or dental arch from an appropriate axial slice, enabling the creation of panoramic views\cite{flores-mir_2014, pittayapat_2013}. However, selecting the optimal axial slice that comprehensively covers each tooth, irrespective of its angulation, demands substantial effort and time. Additionally, accurately identifying critical craniofacial landmarks essential for depicting jaw shape requires the expertise of a skilled professional.

Automated techniques for panoramic synthesis from CBCT primarily delineated a suitable dental arch resembling the jaw shape using appropriate curve fitting algorithms \cite{tohnak_2007, yun_2019, amorim_2020, PS.Anusree_2023}. Depending on the chosen reconstruction methodologies, the reformatted X-ray images can minimize interference from the ghost effect and unwanted anatomical structures. Tohnak et al. introduced an algorithm for reconstructing synthetic X-ray images from CT data, requiring users to mask the maximum intensity projection of CT data in the axial plane or manually delineate a 3D region of interest around the teeth that appear in the final X-ray projection. Yun et al. proposed a panorama synthesis algorithm that involves jaw detection through thresholding axial slices encompassing the teeth and fitting polynomial curves to approximate a dental arch. Alternatively, Amorim et al. utilized Bezier curve fitting to represent the jaw shape. Nonetheless, selecting an optimal curve for defining an ideal dental arch resembling the jaw shape remains an issue. The task becomes intricate when attempting to extract a proper dental arch in the presence of significant metal implants. Given that most approaches prioritize the selection of axial slices with visible teeth, the same methodology proves ineffective for patients without teeth. While manual delineation of the arch is considered a gold standard, it necessitates expert intervention. Accurately estimating the focal plane that encompasses the entire teeth region without compromising the linear and angular measurements \cite{abdinian_2016} of the teeth is another challenge. If the patient's jaws fall outside the focal plane, the image's quality and measurements may degrade\cite{ramakrishna_pawar_2014}, impacting the overall utility of the image. Additionally, some patients may exhibit teeth with incorrect axial inclinations, particularly at the incisors, resulting in unclear visibility in the panoramic image.

 Contemporary panoramic systems integrate movable rotation centers, tracing an elliptical path during jaw scanning \cite{sanderink_1987, welander_1990}. These rotation patterns, tailored by manufacturers, can be adjusted based on the patient's jaw morphology. However, standard trajectories may not consistently yield optimal results due to variations in patient positioning during CBCT acquisition and diverse jaw shapes. So, researchers have also been focusing on replicating panoramic imaging rotational principles using CBCT. Park {\it et al.} \cite{park_2023} heuristically defined a quadratic curve for the dynamic rotation trajectory, however, this fixed trajectory path applies universally, disregarding individual patient jaw shapes and positions. Moreover, the angular shift between the adjacent rays defined as per their approach causes the majority of rays to be centered near the molars. Alternatively, Kwon et al.\cite{kwon_2023} introduced a technique for panoramic image reconstruction by isolating panoramic projection data from dental CBCT projection data. However, the effectiveness of this approach is contingent upon having access to both the CBCT and panorama acquisition parameters. These parameters are essential for accurately extracting the projections from CBCT, ensuring a comprehensive panoramic view without information loss.
 
To address these constraints, our paper introduces an innovative approach to panorama tomosynthesis. Our method involves the definition of a simulated projection geometry that integrates moving rotation centers following a semi-elliptical trajectory. Unlike traditional methods that rely on patient-specific dental arch delineation, our proposed technique aims to identify an optimal focal trough encompassing the entire jaw region, ensuring its inclusion in the final 2D projection. Additionally, our algorithm incorporates a tilt correction in the sagittal plane to mitigate unequal magnification of left and right jaws resulting from improper patient positioning. Notably, our method demonstrates efficacy even in scenarios involving extensive metal implants or patients with missing or non-existent teeth. Our study is pioneering in harnessing an extensive array of datasets to evaluate the efficacy of tasks related to this domain comprehensively. 

The subsequent sections of this paper are structured as follows: Section \ref{methods} outlines the data collection and a comprehensive overview of the steps undertaken in our proposed approach. Our experimental findings and analysis are reported in section \ref{results}. Subsequently, sections \ref{discussions} and \ref{conclusion} provide in-depth discussions and conclusions, respectively.

\section{Materials and Methods}
\label{methods}
\begin{figure*}[htbp]
\centerline{\includegraphics[width = \textwidth]{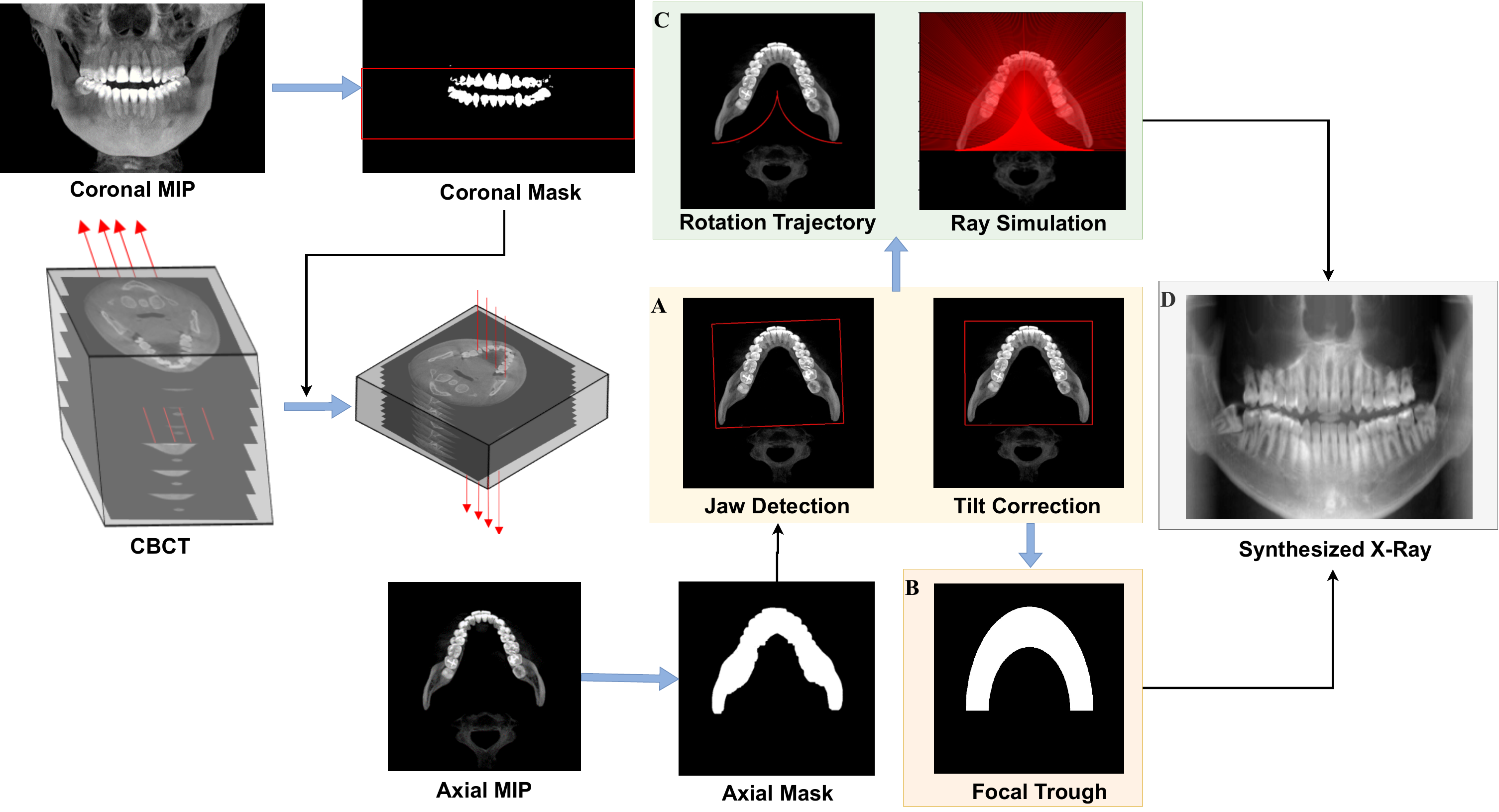}}
\caption{Complete workflow of the proposed algorithm: Coronal and Axial MIPs(Maximum Intensity Projection) images are generated to guide jaw detection. A. Perform Jaw contouring and horizontal tilt-correction. B. Define an elliptical focal trough region. C. Formulate dynamic rotation trajectory and Simulate pencil beams. D. Synthesized Panoramic X-ray. }
\label{fig1}
\end{figure*}

The proposed methodology is outlined in Fig.~\ref{fig1}. It consists of four major tasks: Initially, the jaw position is detected and any misalignment stemming from the patient's head tilt relative to the sagittal reference plane is corrected.  Subsequently, a focal trough is defined, encompassing both the maxilla and mandible, thereby ensuring that only tissues within this delineated region contribute to the resultant Panoramic projection. This measure effectively mitigates the occurrence of unwanted anatomical structures and ghost images in the X-ray output. Following this, a trajectory is determined to replicate the movement of the source-detector system and X-ray directions. Multiple X-ray projections are then generated along this trajectory, covering a 90-degree angular sweep using pencil-beam geometry. The summation of these generated projections yields the final Panoramic X-ray image.

\subsection{Jaw Detection and Tilt-Correction}

A patient's head CBCT encompasses a diverse array of tissues, including air, soft tissues, bone, and enamel, among others. The grayscale values depicted in CBCT scans are not direct representations of Hounsfield Units (HU) but are often linearly scaled values, contributed by various factors such as X-ray tube parameters, cone-beam geometry, detector characteristics, scattering, and other artifacts \cite{pauwels_2013, razi_2014}. To enhance contrast and facilitate tissue differentiation, We have applied rescaling and windowing techniques to the CBCT data according to the manufacturer-defined parameters.  Given the predominant composition of bone and enamel within the jaw structure, extraneous tissues are eliminated by adjusting the window level to encompass values surpassing those of soft tissue.

The Coronal Maximum Intensity Projection(MIP) is obtained from the windowed CBCT data to determine the Region of Interest(ROI) composed of teeth and jaw. ROI is determined in such a way that it encompasses both the mandible and maxilla. Subsequently, an axial MIP is taken using a subset of slices belonging to this ROI region,  providing a preliminary approximation of the jaw shape. However, the presence of metallic implants can disrupt this process, necessitating further refinement of the jaw mask through morphological operations like opening, closing, and hole-filling. Finally, a contour detection algorithm is used to extract the jaw shape. 

Potential errors in patient positioning caused by the tilt from the sagittal reference planes can be identified and corrected with the help of contour detection plots. Correcting misalignments is imperative, as incorrect jaw alignment can introduce asymmetries in the magnification of the jaw structure, which could potentially obscure diagnostic accuracy. 
 
\subsection{Focal Trough Estimation}

There is no standard shape that defines the shape of a human's jaw. Jaw size can be small, medium, or large, and jaw shape can be V-shaped, standard, or U-shaped for diverse individuals based on their age, race, and gender. Manufacturers typically offer a selection of horseshoe-shaped focal troughs for patient imaging, each tailored to accommodate varying jaw sizes and shapes. Notably, these focal troughs exhibit thickness variations, being thicker near the molars and thinner near the incisors.

However, in our approach, we have implemented elliptical focal troughs to encompass the entire mandible and maxilla. These troughs are strategically designed with thicker regions near the incisors and thinner regions near the molars. This adjustment addresses potential challenges, such as anterior displacement of the maxillary anterior region  \cite{martins_2022}, which may obscure the visibility of the maxillary incisors in X-ray images and lead to distortion or misinterpretation of the image. By confining the X-ray projection to the regions enclosed within the focal troughs, we ensure that unwanted anatomical structures and ghost images are minimized in the resulting X-ray outputs. This approach enhances diagnostic clarity and accuracy by reducing the potential for image artifacts and superimposition of irrelevant structures.

\subsection{Panoramic Scan Simulation}

Unlike CBCT machines, Panoramic systems utilize a dynamic rotation center for the movement of the source-detector unit. This movement defines the focal layer within the patient's jaw.  
However, like the dental arches a standard trajectory doesn't exist and varies depending on the manufacturers.  To address this variability and simulate projection geometries tailored to the patient's jaw size, we have developed an elliptical trajectory model, as expressed in \eqref{eq1}.
\begin{align}
 \frac{(x-h)^2}{a^2} + \frac{(y-k)^2}{b^2} = 1 \label{eq1}
\end{align}
where, parameters $h, k, a,$ and $b$ are determined based on the dimensions of the jaw region identified using contour plots. In our method, $b$ is half the height and $a$ is half the width of the contour plot. This approach allows for the formulation of trajectories that adapt to the specific anatomical characteristics of the patient's jaw. A representative trajectory formulated using this methodology is illustrated in Figure .~\ref{fig1}C.

Once the trajectory is established, the next step involves tracing rays along this trajectory. These rays are defined as tangents to the elliptical trajectory with predefined slopes. The equation of a line is given by \eqref{eq2}.
\begin{align}
 y = mx + c \label{eq2}
\end{align}
where m is the slope of the line and c is the y-intercept. 

By combining \eqref{eq1} and  \eqref{eq2}, the slope of the tangent line (m) can be determined, as shown in Eqn. \eqref{eq3}.
\begin{align}
 m = \frac{-b^2(x-h)}{a^2(y-k)} 
 \label{eq3}
\end{align}

To ensure comprehensive sampling of rays throughout the jaw, we adopt a 180-degree total angular sweep and the shift between adjacent rays changing between 0.4 and 0.8. Consequently, the number of sampled rays decreases as the tangents move anteriorly. This approach guarantees uniform ray distribution throughout the jaw, thereby enhancing the accuracy and reliability of panoramic imaging.

Following the extraction of rays, we sample points along these rays, with these points constrained to the predefined focal trough. Subsequently, these sampled points are aggregated into single pixels using Beer Lambert's equation, facilitating the synthesis of X-ray images. This methodology allows for the accurate representation of anatomical structures within the focal trough, thereby contributing to the diagnostic utility of panoramic radiography.

\subsection{X-ray Synthesis}

According to Beer Lambert's absorption-only attenuation law , the resultant intensity I of a pencil beam passing through a material of thickness t and density $\rho$, with linear attenuation coefficient $\mu$ and mass attenuation coefficient $\left[\dfrac{\mu}{\rho} \right]$, can be defined as follows:
\begin{align}
 I = I_{0} \exp \left[-\dfrac{\mu}{\rho} \rho t \right] = I_{0} \exp[-\mu t].
 \label{eq4}
\end{align}

 For a ray of length $l$, this equation can be expressed in integral form as in Eqn.\eqref{eq5}.
 
\begin{align}
 I = I_{0} \, \exp \left[- \int_{0}^{l}\mu (t) dt \right].
 \label{eq5}
\end{align}

By discretizing it into N sampling points along the ray, Eqn. \eqref{eq5} can be reformulated into Eqn. \eqref{eq6}.

\begin{align}
 I \approx I_{0}  \exp \left[ - \sum_{i = 1}^{N} \mu_{i} \epsilon_{i} \right].
 \label{eq6}
\end{align}

We assume that attenuation is independent of the ray direction. Considering the linear relationship between $\mu$, HU, and CBCT gray values, the transmittance of the ray through the CBCT volume \cite{park_2023} can be expressed as:

\begin{align}
 T =  \exp \left[- \sum_{i = 1}^{N} \beta \, \sigma_{i} \,  \delta_{i} \right],
 \label{eq7}
\end{align}

where, $\beta$ is the correction factor heuristically determined from experiments and $\sigma_i$ is the pixel intensity and $\delta_i$ is the unit distance in CBCT domain.

Furthermore, in our process, we have utilized CBCT windowing to suppress tissues below the soft tissue window when rendering the rays. This technique aids in enhancing the visibility of incisors, which might otherwise be obscured due to the area between the tongue and teeth. Ultimately, the rendered value at the pixel location corresponds to $(1-T)$.

\subsection{Dataset}
The clinical dental head CBCT scans of $600$ patients ($400$ Females, $200$ Males; Mean age:$50$, Deviation: $26$) gathered from the Chosun School of Dentistry in Gwangju, South Korea, were used in the study. The dataset, provided in raw DICOM format, consisted of a series of $16$-bit grayscale images with resolutions spanning from $0.25$ to $0.5$ mm. The imaging procedures utilized the CS900 scanner from Carestream Health and the Planmeca VISOG7 scanners. Real panoramic X-rays corresponding to each patient were also available for comprehensive quality assessment.

\section{Results}
\label{results}
\begin{figure*}[htbp]
\centerline{\includegraphics[width = 0.6 \textwidth]{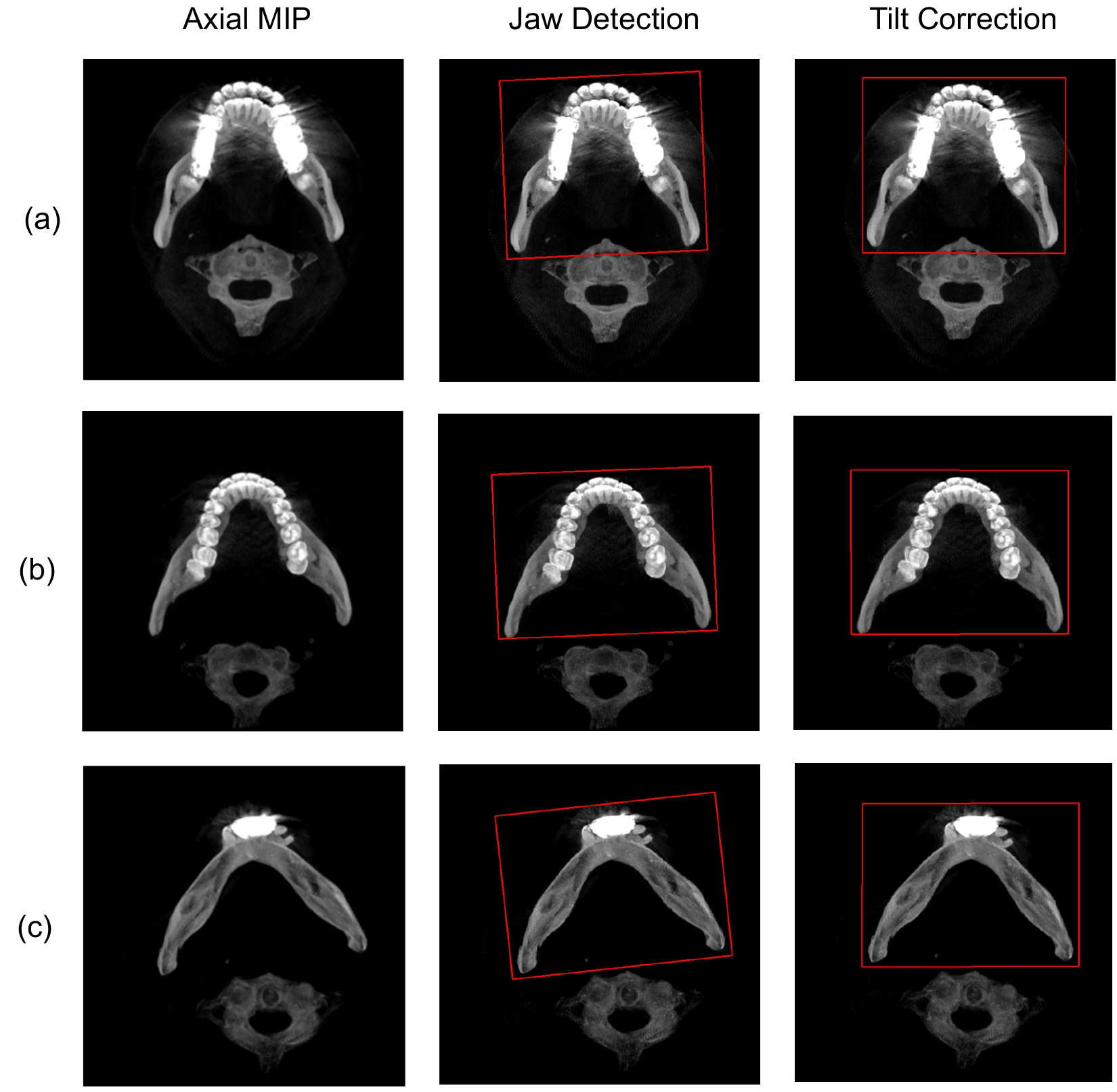}}
\caption{ (a) In the presence of heavy metal implants. (b) Normal jaw (c) Missing teeth }
\label{fig2}
\end{figure*}

Our algorithm underwent successful validation across 600 datasets, demonstrating robust outcomes in jaw detection and tilt correction. Figure. ~\ref{fig2} depicts jaw detection and tilt correction results for three cases. Notably, the algorithm exhibited resilience even with complex factors such as heavy metal implants or missing teeth, reaffirming its suitability for real-world clinical settings. A pivotal aspect of our approach lies in the meticulous preprocessing steps employed to enhance the quality of input data. Leveraging the CBCT windowing techniques, we successfully standardized gray-level variations across the dataset, optimizing tissue differentiation and ensuring consistent image quality.

\begin{figure*}[htbp]
\centerline{\includegraphics[width = \textwidth]{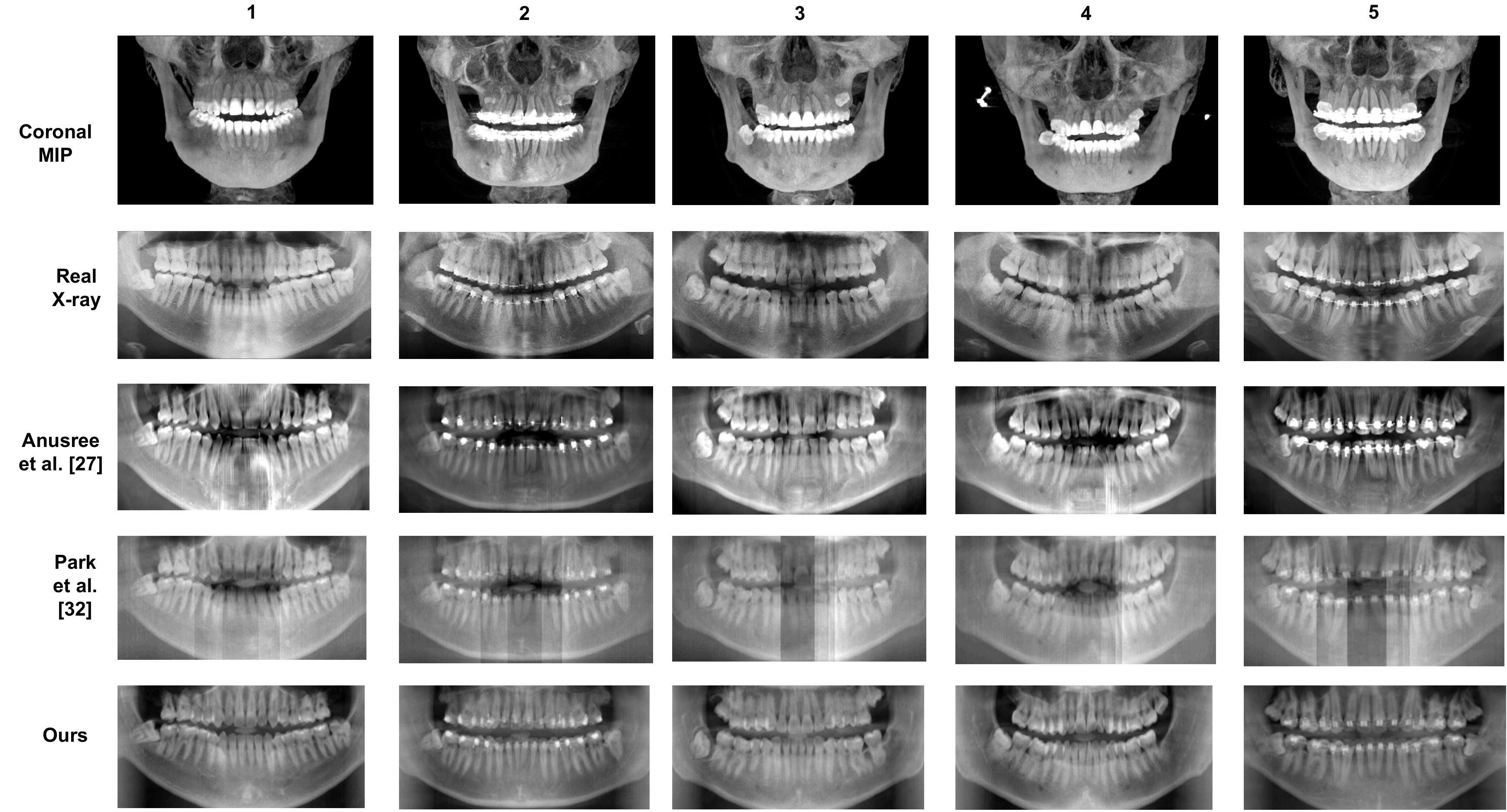}}
\caption{Qualitative comparison for Panorama synthesis results from 3 different approaches.}
\label{fig3}
\end{figure*}

Figure ~\ref{fig3} showcases the qualitative outcomes obtained from the synthesis of panoramic X-rays using our proposed methodology. We have considered two baseline models for comparison. The first baseline model, as described in \cite{PS.Anusree_2023}, employs a method of synthesizing panoramic X-rays by delineating a dental arch and subsequently flattening the CBCT volume along this extracted arch. Conversely, the second baseline model \cite{park_2023}, adopted a similar approach to ours by using principles of rotational panoramic radiography. 

The limitations of the CBCT flattening approach are apparent from the distorted representation of incisors observed in the resulting X-rays. Such distortions stem from the challenges associated with selecting an appropriate dental arch and determining the depth of the flattened slices. Additionally, the anterior tilt of maxillary incisors further exacerbates this issue, often resulting in blurred or entirely invisible incisors in the final X-ray images. In contrast, the approach proposed by Park {\it et al.} \cite{park_2023} introduces inconsistencies in ray sampling near the incisors due to a sudden increase in the sampling rate of rays in this region. Consequently, smaller incisors may appear merged in the final X-ray images, compromising their clarity and distinctiveness.  However, our approach circumvents these imperfections through the implementation of patient-specific rotation trajectories and a consistent increase in shift angles between the extracted rays. This ensures a more uniform and accurate representation of anatomical structures, particularly near the incisors.

To quantitatively evaluate the efficacy of our approach, we conducted a comprehensive analysis on a random set of 10 patients as shown in Table \ref{Tab1}. The results were compared against the baseline models using two content-invariant metrics: Learned Perceptual Image Patch Similarity (LPIPS) and Structural Similarity (SSIM). LPIPS measures the perceptual similarity between images based on learned representations, while SSIM quantifies the similarity of structural patterns within images.

It is essential to note that while LPIPS and SSIM provide valuable insights into the comparative performance of different synthesis methods, they are not absolute measures of accuracy. Variations in the cropping of regions from the original X-ray images can influence the computed metric values. Nonetheless, the findings from our evaluation lend support to the superiority of our approach over the baseline models in producing higher-quality panoramic X-ray images.

\begin{table*}[htbp]
{\centering
\renewcommand{\arraystretch}{1.5}%
\begin{tabular}{|c||c|c|c||c|c|c|}
\hline
\textbf{Dataset} & \multicolumn{3}{c||}{\textbf{LPIPS $(\downarrow)$}} & \multicolumn{3}{c|}{\textbf{SSIM $(\uparrow)$}}\\
\cline{2-7}
 & \textbf{ Ours } & \textbf{Anusree {\it et al.} \cite{PS.Anusree_2023}} & \textbf{ Park {\it et al.} \cite{park_2023}} & \textbf{ Ours } & \textbf{Anusree {\it et al.} \cite{PS.Anusree_2023}} & \textbf{ Park {\it et al.} \cite{park_2023}} \\ 
 \hline
 \hline
Dataset 1 & 0.2193  & 0.2718 & 0.2023 & 0.249 & 0.232 & 0.369 \\
Dataset 2 & 0.2430 & 0.2462 & 0.3155 & 0.327 & 0.214 & 0.251 \\
Dataset 3 & 0.2927  & 0.3047 & 0.3068 & 0.334 & 0.139 & 0.314 \\
Dataset 4 & 0.2867  & 0.2666 & 0.3093 & 0.333 & 0.288 & 0.239 \\
Dataset 5 & 0.2869  & 0.2845 & 0.2893 & 0.231 & 0.246 & 0.264 \\
Dataset 6 & 0.2503  & 0.2571 & 0.2852 & 0.348 & 0.275 & 0.300 \\
Dataset 7 & 0.3198  & 0.3272 & 0.3285 & 0.252 & 0.190 & 0.217 \\
Dataset 8 & 0.3015  & 0.3042 & 0.3055 & 0.222 & 0.220 & 0.237 \\
Dataset 9 & 0.2756  & 0.3035 & 0.3154 & 0.234 & 0.215 & 0.174 \\
Dataset 10 & 0.2722  & 0.3028 & 0.2804 & 0.198 & 0.221 & 0.218 \\
\hline
\hline
Average & \textbf{0.2748}  & 0.2869 & 0.2938 & \textbf{0.273} & 0.224 & 0.258 \\
\hline
\end{tabular}
\caption{Quantitative analysis results for panorama synthesis on 10 random datasets using 3 different approaches.}
\label{Tab1}}
\end{table*}

\section{Discussions}
\label{discussions}

\subsection{CBCT Preprocessing}

CBCT windowing techniques have been strategically employed at various stages of the process, utilizing an approximation of the soft-tissue window within the range of $[-125,225]$. Specifically, during preprocessing, a windowing operation spanning $[225,3096]$ has been implemented to adjust both low intensities, such as air and soft tissue, and high intensities, such as metallic implants. The minimum window level of $-175$ has been set during the X-ray synthesis phase.

The utilization of maximum intensity projections (MIP) holds significant importance in the identification of the ROI containing the teeth and jaw. This process enhances the visualization of structures with high-intensity values along the chosen direction, effectively highlighting features of interest. Gaussian curve fitting is applied to the intensity histogram plot of the MIP image and the threshold for the teeth mask is obtained using Eqn. \eqref{eq7}. From the Coronal MIP mask, the ROI can be identified by fitting another Gaussian curve on its horizontal projection histogram, enabling the determination of the lower and upper limits for the ROI slices. These limits, represented by the values $a$ and $b$ and computed using equation \eqref{eq8}, serve to precisely delineate the extent of the ROI containing the teeth and jaw.

\begin{align}
     \text{Threshold}, t =  \mu + 2 \sigma,
 \label{eq8}
\end{align}
\begin{align}
 a =  \mu - 2.5 \sigma, \\
 b =  \mu + 1.5 \sigma,
 \label{eq9}
\end{align}

where $\mu$ and $\sigma$ are the mean and standard deviation of the Gaussian fit.

\begin{figure*}[htbp]
\centerline{\includegraphics[width = 0.97 \textwidth]{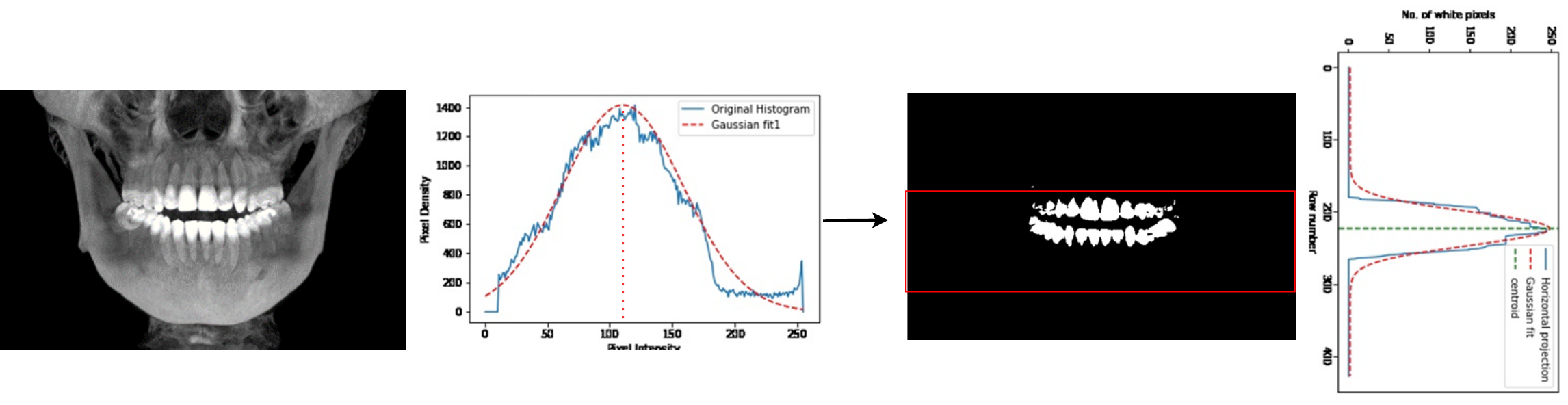}}
\caption{Maximal Intensity Projections and the histogram plots}
\label{fig4}
\end{figure*}

\subsection{Conventional v/s Synthesized X-rays}

Conventional Panoramic systems operate by reconstructing a comprehensive two-dimensional representation of the maxillofacial region through the acquisition of projections from varying angles, facilitated by synchronized movement of the X-ray source and detector along a shifting trajectory. The quality of such X-rays is significantly influenced by a multitude of factors including imaging protocols, equipment settings, and patient positioning. Consequently, images captured by distinct panoramic systems inherently exhibit unique characteristics, precluding exact replication. Evaluation of X-ray quality encompasses critical considerations such as the clarity and visibility of anatomical structures, magnification rate, and the curvature of the smile line. 

For optimal diagnostic utility, the occlusal plane should manifest as either straight or slightly curved upwards \cite{imajo_2024}, albeit variations may arise owing to tooth alignment and patient positioning nuances. Furthermore, diagnostic clarity necessitates the discernible presence of dental fixtures such as braces, prosthetic crowns, restorations/fillings, implants, root canals, and temporomandibular canals. Notably, the identification of inferior alveolar nerves traversing the mandibular canal serves as a valuable indicator for distinguishing radio-opaque and radiolucent regions, thereby facilitating the detection of potential tumors and cysts in this anatomical area. Our synthesized images successfully captured these intricate details following the reconstruction process.

\begin{figure*}[htbp]
\centerline{\includegraphics[width = 0.7 \textwidth]{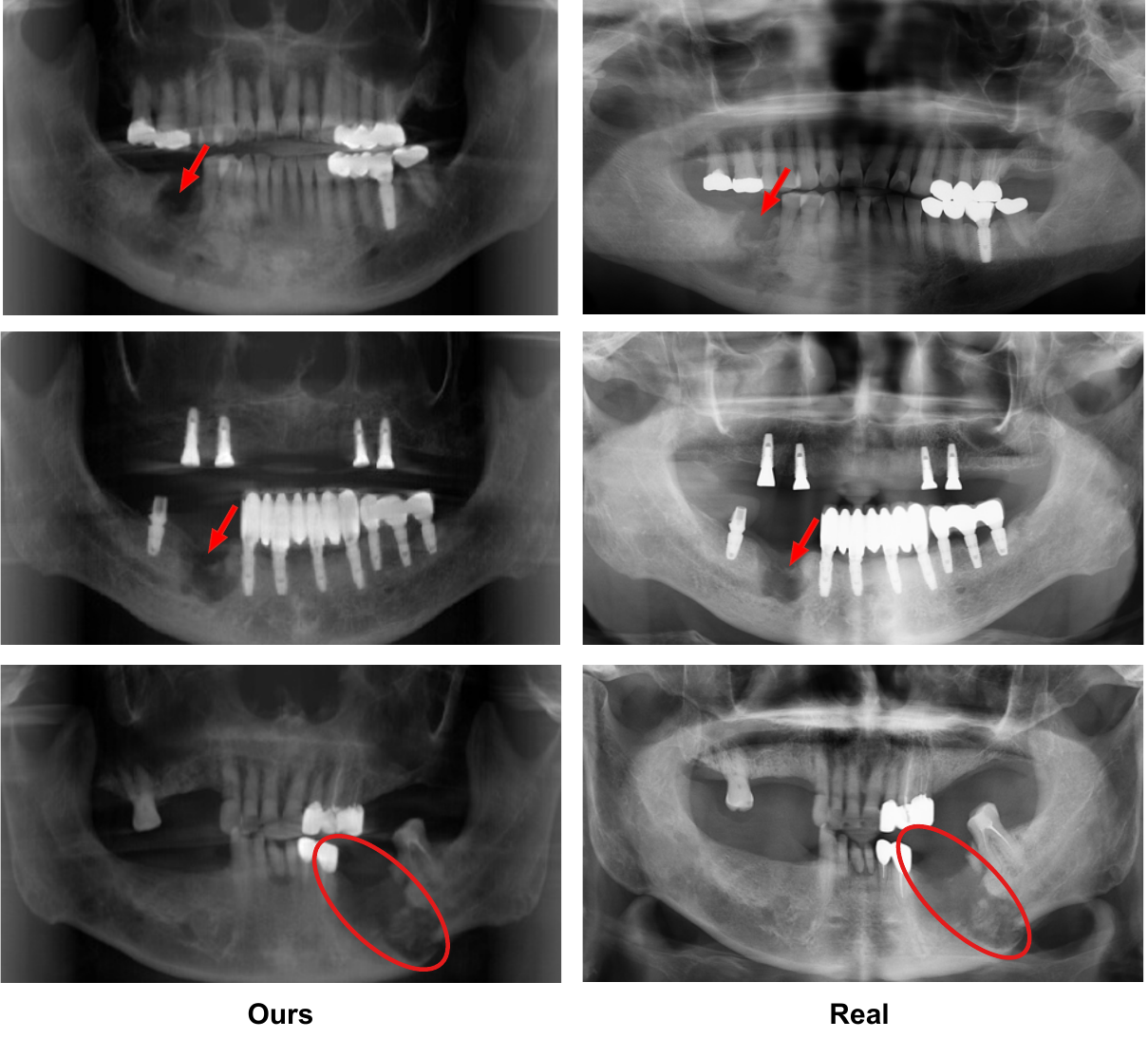}}
\caption{Images with visible bone loss or severe bone damage}
\label{fig5}
\end{figure*}

Panoramic X-rays offer a valuable means of detecting significant bone destruction and loss, which is crucial for assessing the adequacy of bone support required to maintain tooth stability. The integrity of the bone support, extending to the tooth crown, is paramount for ensuring sufficient support for the tooth structure. However, bone resorption can occur due to various patient-related conditions and local factors, leading to a reduction in bone volume and support. Panoramic X-rays serve as a diagnostic tool for quantifying both vertical and horizontal bone losses, aiding in the assessment of overall bone health and the extent of periodontal disease. Figure \ref{fig5} shows that our synthesized X-rays demonstrate the capability to visualize bone loss with a fidelity comparable to that of conventional X-rays. This underscores the significance of our synthesized X-rays as a valuable diagnostic tool for identifying and assessing bone loss, thereby facilitating comprehensive evaluation and management of periodontal conditions and related dental pathologies.

\begin{figure*}[htbp]
\centerline{\includegraphics[width = \textwidth]{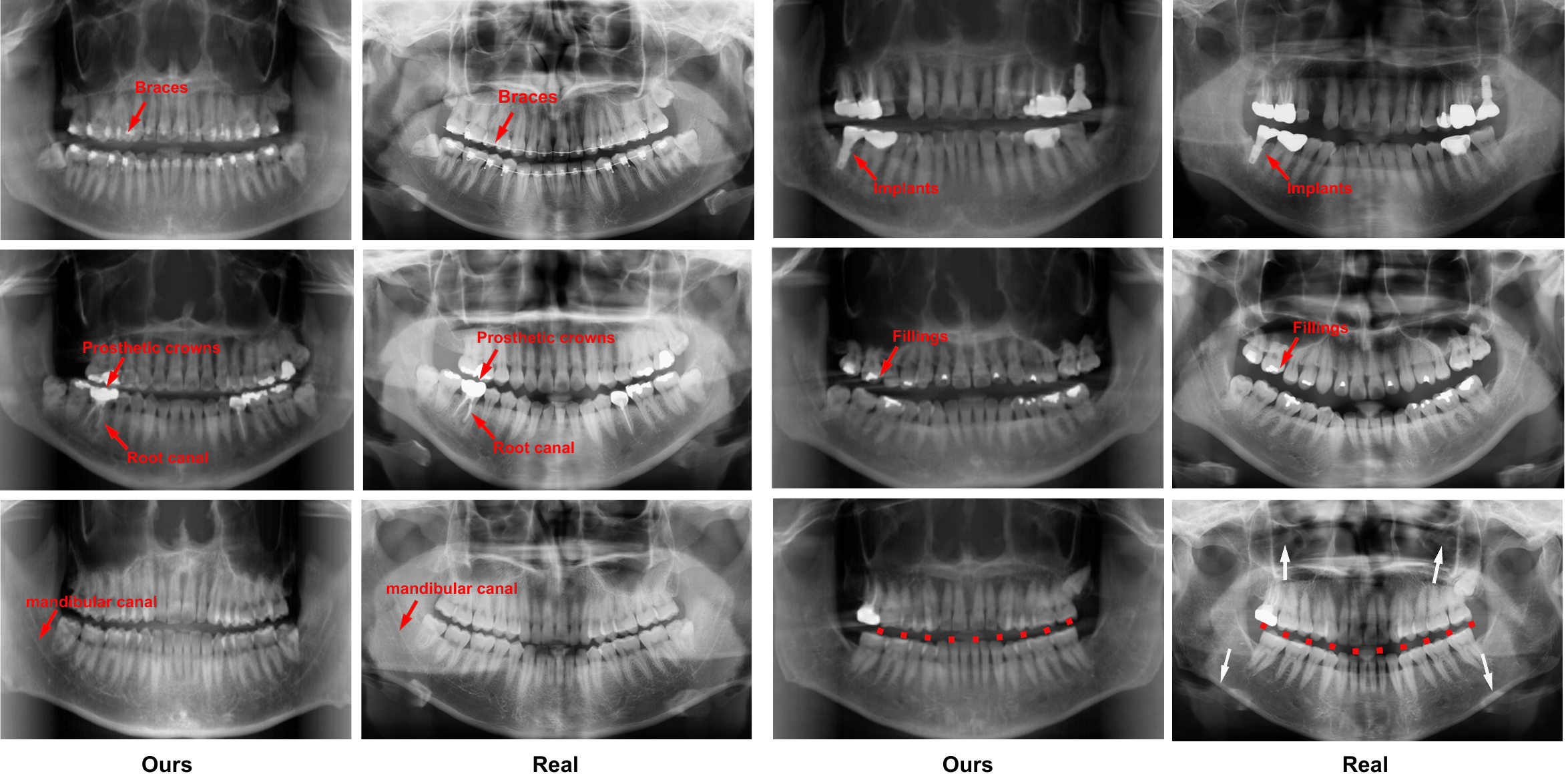}}
\caption{Comparison between real and synthesized X-rays. Some of the disturbances from adjacent structures are highlighted with white arrows in the real X-ray. Such disturbances are not present in the synthesized ones. The dotted red line represents the smile curve.}
\label{fig6}
\end{figure*}

A notable drawback associated with panoramic X-rays lies in the occurrence of ghost structures and the superimposition of images from the opposite jaw and cervical spine, potentially compromising diagnostic accuracy. However, our proposed methodology addresses this limitation by incorporating jaw detection and focal trough estimation techniques, thereby ensuring that only structures within the predefined region of interest are projected onto the final image. Consequently, this approach effectively mitigates disruptions originating from neighboring structures, thereby augmenting the overall diagnostic efficacy of the panoramic X-rays. A comparison between real X-rays and the corresponding synthesized X-rays is shown in Fig. \ref{fig6}.

\subsection{Limitations}

Despite the considerable advantages offered by our synthesized X-rays, it's essential to acknowledge their inherent limitations. For an ideal Panoramic X-ray, the focal trough should conform precisely to the shape of the patient's jaw. It should be sufficiently generous enough to encompass the entirety of the teeth and jaw, both anteriorly and posteriorly, to ensure comprehensive visibility of incisors and molars. Hence, the width of the focal trough shouldn't remain uniform across all axial slices, and each axial focal plane needs to be anteriorly or posteriorly displaced to encompass the necessary structures. This dynamic adjustment is crucial to ensure the X-ray maintains the requisite sharpness and dimensional accuracy for accurate diagnosis and effective treatment planning. As noted by Scarfe {\it et al.} \cite{scarfe_1993, scarfe_1998}, optimal beam projection angles are pivotal in minimizing the overlap between teeth, particularly in regions prone to inter-proximal overlap, such as the premolars. Hence, contrary to earlier suggestions, the optimal projection angle for the rays should not be 90 degrees to the dental arch. Achieving precise alignment with the diverse inter-proximal contact points is challenging due to dental deformities in the patient. The individual's gender, age, and race also influence this. Hence, our choice of shape for the rotation trajectory cannot adequately handle this requirement due to patient diversity.

While our paper has effectively tackled the issue of patient positioning errors resulting from the tilt from the sagittal reference plane, we have yet to address similar concerns regarding the axial and coronal reference planes. Identifying suitable landmarks to ascertain the tilt from these reference planes necessitates expert intervention, potentially requiring manual adjustment. So, further research and development are warranted to address similar challenges concerning the other two planes. Additionally, the presence of scattering from metal implants and other artifacts originating from CBCT reconstruction can significantly compromise the quality of synthesized X-rays. Although this aspect has not been specifically addressed in our current paper, we aim to further enhance the accuracy and reliability of our synthesized X-ray imaging technique in our future work.

\section{Conclusion}
\label{conclusion}
In this paper, we have discussed a novel approach for panorama synthesis from CBCT leveraging the rotational principles of panoramic radiography using a simulated projection geometry with moving rotation centers. Our work has systematically tackled numerous challenges associated with image synthesis, such as the presence of metal artifacts, incorrect patient positioning, and absence of teeth. The efficacy of our proposed approach was rigorously evaluated across a diverse dataset sourced from two different CBCT machines. The empirical findings from our experiments attest to the method's proficiency in faithfully recovering intricate geometric details of teeth, thereby providing a valuable contribution to the field of panoramic imaging synthesis from CBCT data.

\section*{Funding}
This work was supported by the Basic Science Research Program (NRF-2022R1F1A1075204), the BK21 FOUR, and the Regional Innovation Strategy Project (2021RIS-004) through the National Research Foundation of Korea (NRF) funded by the Ministry of Education, South Korea.

\section*{Data Availability}
The datasets used in this study are not available publicly but will be available from the corresponding author upon reasonable request.

% \section*{Competing Interests}
% The authors declare no competing interests.

\section*{Additional Information}
\textbf{Correspondence} and request for materials or dataset should be addressed to W. You.

%Bibliography
\bibliographystyle{unsrt}  
\bibliography{main}

\end{document}